# Bayesian sample size determination of vibration signals in machine learning approach to fault diagnosis of roller bearings

**Siddhant Sahu[*], V. Sugumaran[**]**

[*] School of Mechanical and Building Sciences, VIT University, Chennai
Email: siddhant.sahu2011@vit.ac.in

[**] School of Mechanical and Building Sciences, VIT University, Chennai
Email: v_sugu@yahoo.com

**ABSTRACT**

Sample size determination for a data set is an important statistical process for analysing the data to an optimum level of accuracy and using minimum computational work. The applications of this process are credible in every domain which deals with large data sets and high computational work. This study uses Bayesian analysis for determination of minimum sample size of vibration signals to be considered for fault diagnosis of a bearing using pre-defined parameters such as the inverse standard probability and the acceptable margin of error. Thus an analytical formula for sample size determination is introduced. The fault diagnosis of the bearing is done using a machine learning approach using an entropy-based J48 algorithm. The following method will help researchers involved in fault diagnosis to determine minimum sample size of data for analysis for a good statistical stability and precision.

## I. INTRODUCTION

Bearings are an integral part of any machinery used in the industry nowadays. It is basically a machine element that enables relative motion of one part with respect to the other thus reducing the friction between any two parts. It enables linear movement of a specific part or rotary motion about a fixed axis. Thus, being such an essential component, if the bearing carries some fault or abnormality, it directly affects the efficiency of the machine and leads to further damage or wear and tear. This leads to a need of fault diagnosis thus ensuring proper working and maintenance of the machine in which it is being used.

This paper focuses on the machine learning approach for evaluating the fault in various sections of a bearing namely inner race and outer race. For this particular data set, four conditions of the bearing that were taken into account namely 'Normal', 'Inner Race Fault', 'Outer Race Fault' and 'Inner Outer Race Fault'.

The machine learning approach uses a classifier which uses various decision tree algorithms to analyse the data, classify the conditions and finally provide the researcher with the results. The classifier requires a specific number of vibration signals to be able to classify the data with an optimum level of accuracy termed as the classifier accuracy. The major question that arises here is that what is the optimum number of vibration signals to be given as an input to the classifier to have an optimum level of classification accuracy for precise factor. A higher level of classifier accuracy makes it robust. Secondly, a larger number of vibrational signals, when given as an input into a classifier would lead to increase in the computational effort. Thus, in order to meet the above stated requirements, an optimum sample size needs to be chosen.

In this paper, a Bayesian approach based method is used to evaluate the optimum sample size from a normally distributed population of a single type of statistical feature of the vibration signal which best represents the vibration signal as inferred by the classifier. This method is based on the Bayesian theory which uses the present or the prior information and the observed data to evaluate the posterior estimates or the estimated or expected data. Certain parameters of the prior and posterior data and used in a combined way to obtain the optimum sample size. This method is however different from the classical/frequentist method [1] which uses only the prior estimates to evaluate the sample size. Bayesian sample size determination has been applied in various fields, to name a few, 'clinical trials [2]', 'Continuous medical tests [3]', 'Stratified random sampling of pond water [4]', 'ANOVA Bayesian point of view [5]'. However, little work has been done for sample size determination in the field of fault diagnosis using machine learning approach. One of the works done is this field is 'Sample size Determination using power analysis [6]'. 'Inspection of general corrosion of process components [7]', 'Case-control studies with misclassification [8]'. The literature survey for this study was found in 'R. Weiss [9]' Adcock [10], De Santis [11]. The formula used in this study depends on the parameters namely sample standard deviation, prior standard deviation, acceptable margin of error, specified confidence level. The formula finally yields a sample size which ensures that the posterior estimate of the data does not exceed a pre-defined acceptable margin of error.

## II. EXPERIMENTAL DATA

The data set which is being considered for analysis in the present study is the same as the one used by Cohen (1988 [12]), Sugumaran, Sabareesh, and Ramachandran (2008 [13]). The details about the experiment that was conducted which include the setup, fault simulation details and experimental procedure have been explained in Sugumaran and Ramachandran (2007 [14]). The various parameters for the experiment are, sampling frequency (12,000 Hz) and sample length of 8192. (Every vibrational signal has a total count of 8192 data points). For this experimental setup, SKF6205 bearing was considered. Hundred vibration signals were taken by simulating four different conditions on the bearings namely 'Normal', 'Inner race fault', 'Outer Race Fault', 'Inner-outer race fault'. Thus, the final data set consists of a total of 400 vibration signals. The transducer used for data acquisition was a piezo-electric type accelerometer.

## III. FEATURE EXTRACTION

### 3.1. Definitions

Each vibration signal is a culmination of 8192 vibration values. Thus, in order to obtain parameters that suitably represent a specific vibration signal and distinguish it from the other, feature extraction is done. They are explained below as:-

(a) Standard Error: It is defined as a measure of the amount of error in the prediction of y for an individual x in the regression, where x and y are the sample means and 'n' is the sample size. Standard error of the predicted Y

$$= \sqrt{\frac{1}{n-2}\left[\sum(y-\bar{y})^2 - \frac{[\sum(x-\bar{x})(y-\bar{y})]^2}{(x-\bar{x})^2}\right]}$$

(b) Standard deviation: The measure of the effective energy or power content of the sound signal. The following formula was used for computation of standard deviation.

$$\text{Standard Deviation} = \sqrt{\frac{\sum x^2 - (\sum x)^2}{n(n-1)}}$$

(c) Sample variance: The measure of spread of the signal points.

$$\text{Simple Variance} = \frac{\sum x^2 - (\sum x)^2}{n(n-1)}$$

(d) Kurtosis: Indicates the measure of flatness or the spikiness of the signal.

$$\text{Kurtosis} = \left\{\frac{n(n+1)}{(n-1)(n-2)(n-3)}\sum\left(\frac{x_i-\bar{x}}{s}\right)^4\right\} - \frac{3(n-1)^2}{(n-2)(n-3)}$$

(e) Skewness: The characterization of the degree of asymmetry of a distribution about its mean.

$$\text{Skewness} = \frac{n}{n-1}\sum\left(\frac{x_i-\bar{x}}{s}\right)^3$$

(f) Range: Difference in global maxima and minima of the signal.

(g) Minimum value: It is that value which is the global minima of the signal.

(h) Maximum value: It is that value which is the global maxima of the signal.

(i) Sum: Sum of all feature values for each sample.

*3.2 Best Feature Selection*

Firstly, all the statistical features stated above were extracted from each vibration signal containing 8192 data points. Since, all the eleven extracted statistical features are not necessary for classification; these statistical features were subjected to dimensionality reduction. It is a process which eliminates the less useful or unwanted statistical features thus reducing the dimension of the data-set and making it more suitable to be given as an input to the classifier. The classification is done using the J48 algorithm whose theory, procedure and implementation are explained in Sugumaran et. al. (2007) [15]. The decision tree dimensionally reduces the statistical features to four most important features. In this study, the four features in the decreasing order of importance were 'Kurtosis', 'Standard Error', 'Range', 'Minimum'. Out of these four features, further classification using J48 algorithm yields the best statistical feature that could represent the population of vibration signals as a whole with high statistically stability and accuracy. This is shown in the fig. 1.

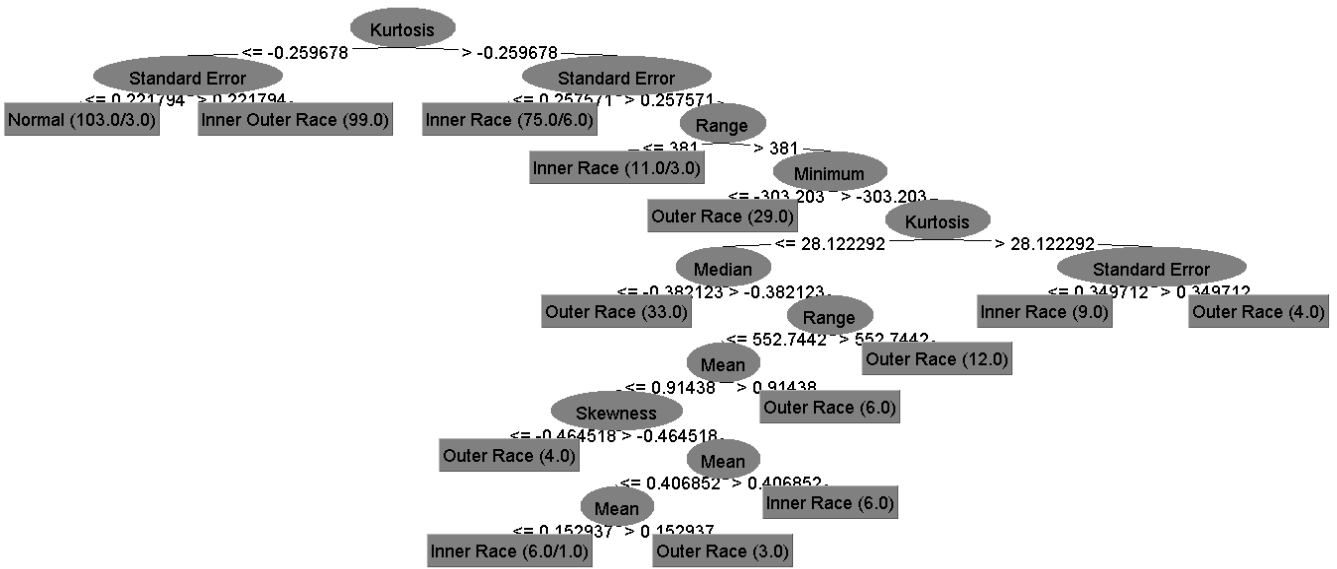

Fig. 1. Decision tree representing the best feature selection process

From fig. 1, it can be inferred that the best statistical feature that represents the data is 'Kurtosis'. Thus, the final data set that was used for analysis was the kurtosis of 400 vibration signals belonging to four different classes, hundred signals in each class.

## IV. BAYESIAN APPROACH BASED METHOD

Bayesian theory is based on the concept of using the prior data to calculate the posterior estimates. The prior information is defined as the information which is currently available or could be based on historical data and theoretical models. Posterior information is defined as the updated information which is obtained using Bayesian updating theory and thus yields a new set of inspection data for analysis. This method stands apart as it used the updated posterior estimates obtained from the prior information whose details and procedures are explained in M. Khalifa, Khan & Haddara (2011) [7].

*4.1 Margins of Error and Finite Population Correction Factor*

Margin of error of mean is the difference between the population mean, μ and the selected sample mean, μ$_s$. mathematically, it is represented as:

$$MOE_{mean} = |\mu - \mu_s| \tag{1}$$

Thus, this parameter determines the accuracy to which the selected sample represents the population as a whole. The lesser the margin of error, the more accurate is the sample, and the greater is the statistical stability. Acceptable margin of error is pre-defined value assigned by the user who is evaluating the sample size. It is one of the two major parameters which decide the optimum sample size using a specific formula. An adequate sample is chosen such that the margin of error does not exceed the pre-defined value at a particular confidence level. In this case, the confidence level signifies the likelihood that the deviation between the population and the sample estimate is not exceeding the margin of error. The margin of error is expressed as:

$$MOE_{accept} = P\left(\frac{1-\alpha}{2}\right) \cdot \frac{\sigma}{\sqrt{n}} \tag{2}$$

Where, 'P' is the inverse standard normal probability at a defined confidence interval α. It is the probability for half the width of the confidence interval for the probability distribution of the mean. The above equation can be rearranged to obtain a formula for the required sample size:

$$n = \left[P\left(\frac{1-\alpha}{2}\right) \cdot \sigma / MOE_{accept}\right] \tag{3}$$

The above formula is only valid if the population size (N) is so large in comparison to the sample size that it can be considered as infinite. However, in order to evaluate the sample size for the case where the population is finite, a correction factor is used which is termed as the finite population correction factor (FPCF) which is expressed as follows: (Bernstein & Bernstein, 1999 [16])

$$FPCF = \sqrt{(N-n)/N-1} = \sqrt{(1-n)/N} \tag{4}$$

The above approximation is done on the basis that the value of N is very large in comparison to 1.

*4.2 Bayes Theorem and Posterior Estimates*

Bayes Theorem states that if $x$ denotes a variable representing a specific data-set, and the distribution of $x$ has a parameter $\beta$, such that the parameter $\beta$ is a random variable with a distribution denoted by $f(\beta)$, the updated or posterior probability distribution, $f''(\beta)$ of this prior distribution can be obtained using Bayesian updating theory using the following formula:

$$f''(\beta) = \frac{P(x|\beta).f(\beta)}{\int_{-\infty}^{\infty} P(x|\beta)f(\beta)d\beta} \tag{5}$$

Where $P(x|\beta)$ is the conditional probability of observing the outcome of $x$ for a given $\beta$. This is also termed as the likelihood function of $\beta$.

In order to evaluate the posterior estimate of the parameter β, the following formula can be used:

$$\beta'' = \int_{-\infty}^{\infty} \beta f''(\beta)d\beta \tag{6}$$

The data set considered in this study is the statistical feature which best represents the whole vibration signal, which in this case is the 'Kurtosis'. Thus the variable $x$ representing the data set in this case is kurtosis. Considering the sample mean of the kurtosis of vibration signals of different classes to be the parameter $\beta$, certain assumptions are taken. Assuming that the prior distribution of the sample $\beta$ is normally distributed, using Bayes Theorem, the posterior sample mean and standard deviation can be evaluated as (Ang & Tang, 2007 [1]):

$$\mu''_{\mu_s} = \frac{\mu_s.(\sigma'_{\mu_s})^2 + \mu'.\left(\frac{\sigma_S^2}{n}\right)}{(\sigma'_{\mu_s})^2 + \left(\frac{\sigma_S^2}{n}\right)} \tag{7}$$

$$\sigma''_{\mu_s} = \sqrt{\frac{(\sigma'_{\mu_s})^2.\left(\frac{\sigma_S^2}{n}\right)}{(\sigma'_{\mu_s})^2 + \left(\frac{\sigma_S^2}{n}\right)}} \tag{8}$$

Where,

$\sigma'_{\mu_s}$ is the prior standard deviation of the sample mean probability distribution.

$\sigma_s$ is the standard deviation of the sample of size n.

$\mu'$ is the prior mean of the population.

As discussed earlier, if from a population of prior standard deviation $\sigma'$, if all the possible samples of size *n* are sampled, then the posterior estimate of standard deviation of the probability distribution of the sample mean, $\sigma'_{\mu_s}$ can be expressed as:

$$\sigma'_{\mu_s} = \sigma'/\sqrt{n} \tag{9}$$

Substituting equation (9) in equations (7) and (8)

$$\mu''_{\mu_s} = \frac{\mu_s \cdot \sigma'^2 + \mu' \sigma_s^2}{\sigma'^2 + \sigma_s^2} \tag{10}$$

$$\sigma''_{\mu_s} = \sqrt{\frac{\left(\frac{\sigma'}{n}\right)^2 \cdot \left(\frac{\sigma_s^2}{n}\right)}{\left(\frac{\sigma'}{n}\right)^2 + \left(\frac{\sigma_s^2}{n}\right)}} \tag{11}$$

$$\sigma''_{\mu_s} = \sqrt{\frac{\sigma'^2 \cdot \sigma_s^2}{\sigma'^2 + \sigma_s^2}}/\sqrt{n} \tag{12}$$

$$\sigma''_{\mu_s} = \sigma''/\sqrt{n} \tag{13}$$

Where $\sigma'' = \sqrt{\frac{\sigma'^2 \cdot \sigma_s^2}{\sigma'^2 + \sigma_s^2}} \tag{14}$

Thus, we can estimate the posterior estimates of sample mean and standard deviation from the prior estimates using equations (10) and (13).

*4.3 Sample Size Calculation*

Thus, the posterior estimates for sample mean and standard deviation can be estimated for a finite population using the finite population correction factor. Thus, combining equation (3) and (13):

$$\sigma''_{\mu_s} = \frac{\sigma''}{\sqrt{n}} \cdot \sqrt{1 - \frac{n}{N}} = \sigma'' \cdot \sqrt{\frac{1}{n} - \frac{1}{N}} \tag{15}$$

Thus, in this case, the margin of error of mean will depend on the inverse standard normal probability at a specific confidence level and the posterior estimate of the standard deviation of the sample mean thus replacing the standard deviation of the prior standard deviation of sample mean.

$$MOE_{accept} = \text{P}\left(\frac{1-\alpha}{2}\right) \cdot \sigma''_{\mu_s} \tag{16}$$

Thus, equation (15) and (16) can be used define a formula to evaluate the required sample size *n* at a pre-defined margin of error.

$$n = \frac{\left[P\left(\frac{1-\alpha}{2}\right) \cdot \left(\frac{\sigma''}{MOE_{accept}}\right)\right]^2}{1 + \frac{\left[P\left(\frac{1-\alpha}{2}\right) \cdot \left(\frac{\sigma''}{MOE_{accept}}\right)\right]^2}{N}} \tag{17}$$

Where $\sigma'' = \sqrt{\frac{\sigma'^2 \cdot \sigma_s^2}{\sigma'^2 + \sigma_s^2}}$

## V. RESULTS AND DISCUSSION

In this study, initially, all the thirteen statistical features were extracted from the vibration signals belonging to different classes. In order to perform dimensionality reduction on the data, initially, all the signals were given as in input to the classifier and the best features was identified from the decision tree obtained after applying the J48 algorithm. For this data set, the best feature was 'Kurtosis'. The kurtosis of each vibration signal was separated from the data containing all the statistical features and finally, a population of 400 kurtosis values of vibration signals was obtained containing 100 values from each of the four conditions namely 'Normal', 'Inner Race Fault', 'Outer Race Fault', 'Inner-Outer Race Fault'. The population mean and standard deviation was computed which were 7.8093 and 19.9251 respectively. Next, a Bayesian approach based method was applied to a multi-class problem where a confidence level of 95 % was considered and different values of pre-defined acceptable margin of error for each class were considered to obtain different sample sizes using the formula in equation (17). The sample size obtained using this formula is the total sample size which is the sum of sample sizes of all the four classes. Since, the population of vibration signals contains equal number of data points from each class, thus the same can be assumed for the chosen sample size as well. The formula in equation (17) was applied using an iterative process wherein, a specific sample size was assumed and then a random sample for the particular sample was generated.. The sample standard deviation, $\sigma_s$ was calculated from this randomly generated sample. If the sample size *n* and the sample standard deviation satisfy the above proposed formula, then the iterative process was stopped and the optimum sample size was achieved. In this study, the random sample was generated, whereas when the experiment is being actually conducted, the vibration signals should be randomly taken. The obtained sample sizes corresponding to different values of pre-defined margin of errors using a confidence level of 95 % are shown in Table 1.

Table 1: Sample sizes of vibration signals for different margin of errors.

| S. No. | Margin of Error | Total Sample Size | Sample Size Per Class |
|---|---|---|---|
| 1. | 0 | 400 | 100 |
| 2. | 0.2 | 394 | 99 |
| 3. | 0.4 | 370 | 93 |
| 4. | 0.6 | 339 | 85 |
| 5. | 0.8 | 308 | 77 |
| 6. | 1 | 272 | 68 |
| 7. | 1.2 | 238 | 60 |
| 8. | 1.4 | 172 | 43 |
| 9. | 1.6 | 148 | 37 |
| 10. | 2 | 99 | 25 |
| 11. | 2.3 | 54 | 14 |
| 12. | 2.7 | 34 | 9 |

*5.1 Variation of sample size with classification accuracy*

The variation of sample size with classification accuracy was studied by reducing 5 data points every time from the class of 100 data points and noting the respective classification accuracy which is shown in fig. 2.

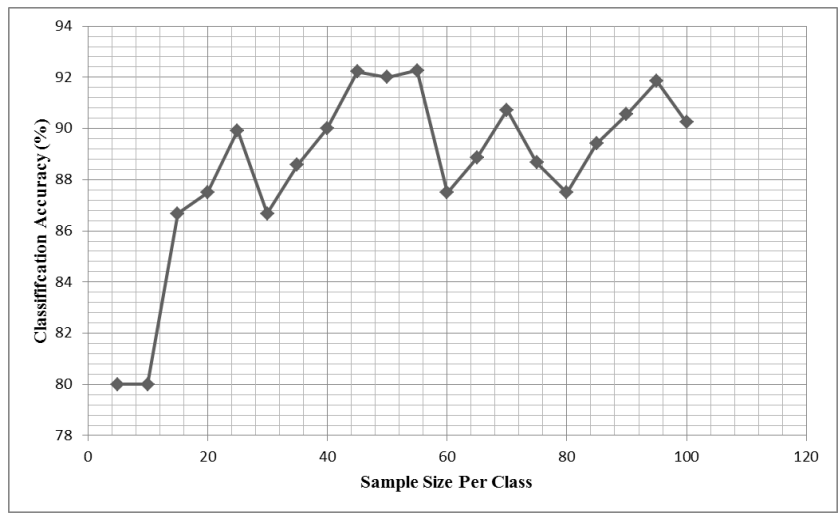

Fig. 2. Classification accuracy as a function of sample size.

From the graph in fig. 2, it can be observed that classification accuracy is stable with less oscillation at sample sizes more than 50 and is less stable at sample sizes less than 50. It falls abruptly to a low value of 80 % below a sample size of 15. For obtaining a sample size, the classification accuracy of a class sample size of 100 will be considered as a reference value.

*5.2 Variation of sample size with error values*

The graphs shown in fig. 3 and 4 depict the variation of the root mean square error and mean absolute error as a function of the sample size by using the same method as mentioned in section *7.1*.

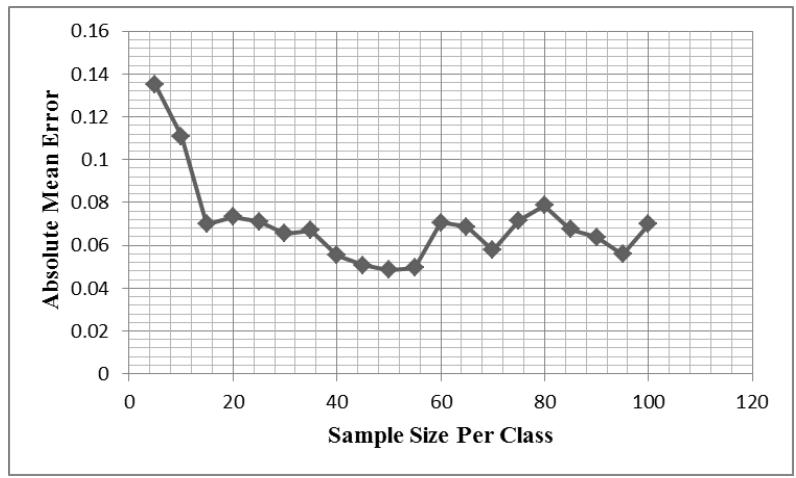

Fig. 3. Variation of absolute mean error with sample size.

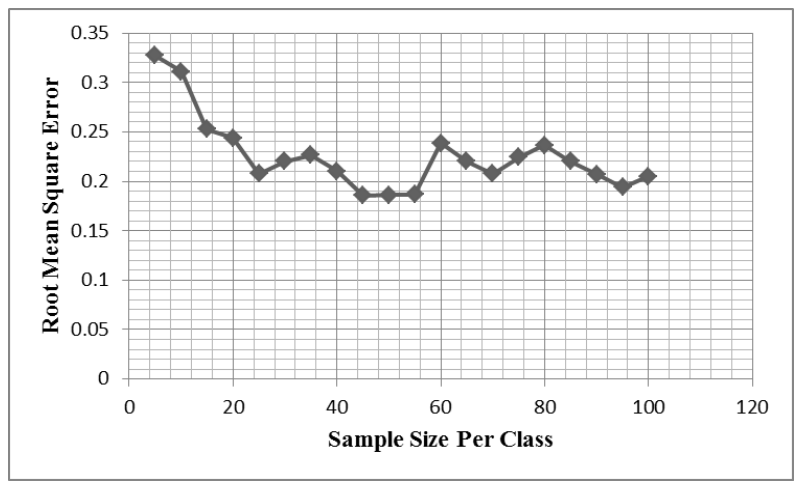

Fig. 4. Variation of root mean square error with sample size.

The two types of errors namely mean absolute error and root mean square error play an important role in the decision of an optimum sample size as these reflect upon the robustness of the classifier. The lesser the error, the better will be the classification accuracy. It can be inferred from fig. 3 and 4 that a sample size higher sample size yields a low absolute mean and root mean square error.. The error values increase drastically as the sample size falls below 20.

*5.3 Optimum Sample Size Determination*

After considering all the arguments discussed in section *7.1, 7.2, 7.3,* and considering the population size to be the reference value, it can be observed that, sample sizes around 45 per class yield a classification a classification accuracy of around 92 % which is prominently high and also yield a low absolute mean and root mean square error. Thus, it could represent a statistically stable sample. However, a sample size of 25 per class yields a classification accuracy of 89.92 %, an absolute mean error of 0.0712 and a root mean square error of 0.2076 which is almost near to the parameters of the actual population which has an accuracy, absolute mean error and root mean square error of 90.25, 0.07, 0.2050. Though, the parameters of this sample size are comparatively lower as compared to those of the sample sizes near 45, but a sample size of 25 per class is a much lower sample size as compared to 45 per class and possesses an almost equally prominent statistically stability as the actual population. Choosing a total sample size of 100 (25 per class) would yield an accurate result and reduce computational effort as well.

*5.4 Variation of sample size with pre-defined acceptable margin of error*

The variations of the sample size with the input parameters in the iterative formula were studied. The variation of sample size with the change in pre-defined acceptable value of margin of error for population of vibration signals for all the four conditions fixed confidence levels (1-*σ*) of 90 %, 95 %, 99%, 99.5%, 99.8% is shown in Fig. 5.

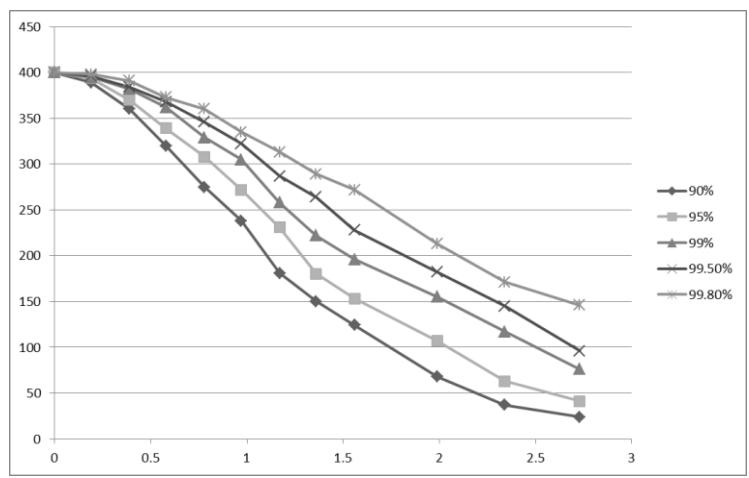

Fig. 5. Variation Of Sample Size With Margin Of Error

From the graph shown in fig. 5, it could be inferred that the sample size increases with a decrease in the pre-defined acceptable value margin of error. Lower margin of error yield a bigger, yet more precise sample size whereas, a higher margin of error yield a smaller, yet less precise sample size. The basic aim is finding a sample size which is satisfactorily precise, and has good classification accuracy as well.

*5.5 Variation of sample size with confidence interval*

The variation of sample size with the change in the inverse standard probability was studied for confidence intervals, (1-$\sigma$) of 90 %, 95 %, 99%, 99.5%, 99.8% keeping the pre-defined acceptable margin of error constant for the population. The evaluated results for population of all the four classes are shown in Fig. 6. The three curves correspond to margin of errors: 0.78, 1.56 and 2.34 respectively.

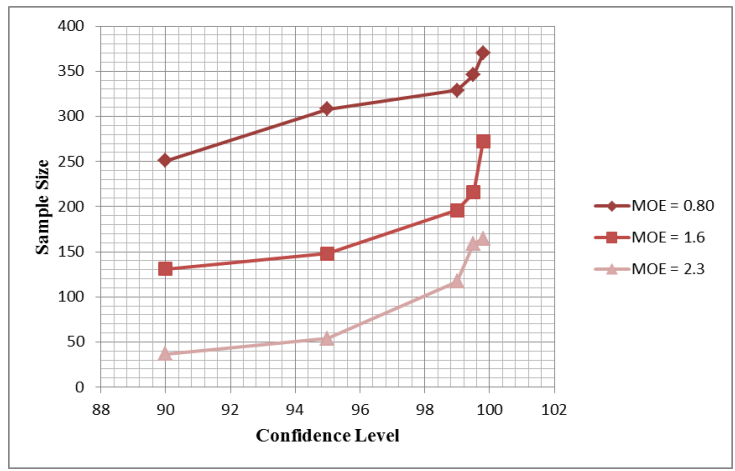

Fig. 6. Variation of sample size with confidence level

From the graph shown in fig. 6, it could be inferred that the sample size increases with the increase in the confidence level. At confidence levels beyond 99 %, the sample size increases drastically. Thus, a larger sample size indicates more likelihood that the deviation between the population and the sample estimate is not exceeding the margin of error. Confidence levels beyond 99 % represent an extremely high level of likelihood and larger sample sizes are required for such high levels.

## VI. CONCLUSION

Thus, the sample size was calculated at a specific confidence level and an acceptable margin of error in the posterior estimate of mean both of which were defined by the user. This particular formula is applicable for a population of size *N*. The variation of the sample size with the pre-defined parameters was discussed. Further, the variation of classification accuracy, root mean square and mean absolute error is also discussed. The classification accuracy and the two types of errors were considered as a basic for choosing the optimum sample size with reference to the respective values of the original population to make the classifier robust and to maintain statistical stability. Finally, a total sample size of 25 per class was chosen as the optimum sample size of vibration signals to be taken for fault diagnosis using machine learning. Thus, the number of vibration signals to be obtained randomly from various conditions of a bearing is reduced from 100 to 25 per class for future experiments.